\documentclass[supp]{new_tlp} 
\usepackage{mathptmx}
\usepackage{tikz}
\usetikzlibrary{arrows, shapes}
\usetikzlibrary{fadings}
\usepackage{abbrev}
\usepackage{amsmath}
\usepackage{amsfonts}
\usepackage{amssymb}
\usepackage{supp}

\definecolor{gold}{rgb}{0.1,0.1,0.1}

 \title{Properties of Stable Model Semantics Extensions}

\author[M\'ario Abrantes \and Lu\'{\i}s Moniz Pereira]
{M\'ario Abrantes\\
Departamento de Matem\'atica, Escola Superior de Tecnologia e de Gest{\~a}o,\\ Instituto Polit\'ecnico de Bragan\c{c}a, 5300-253 Bragan\c{c}a, Portugal\\
\and Lu\'{\i}s Moniz Pereira\\
Centro de Intelig\^encia Artificial (CENTRIA), Departamento de Inform\'atica,\\ Faculdade de Ci\^encias e Tecnologia, Universidade Nova de Lisboa, 2829-516 Caparica, Portugal}

\pagerange{\pageref{firstpage}--\pageref{lastpage}}
\volume{\textbf{10} (3)}
\jdate{March 2002}
\pubyear{2002}

\begin{document}


\maketitle

 \begin{abstract}
The stable model (SM) semantics lacks the properties of existence, relevance and cumulativity. If we prospectively consider the class of conservative extensions of the SM semantics (i.e., semantics that for each normal logic program P retrieve a superset of the set of stable models of P), one may wander how do the semantics of this class behave in what concerns the aforementioned properties. That is the type of issue dealt with in this paper. We define a large class of conservative extensions of the SM semantics, dubbed affix stable model semantics (ASM), and study the above referred properties into two non-disjoint subfamilies of the class ASM, here dubbed $ASM^h$ and $ASM^m$. From this study a number of results stem which facilitate the assessment of semantics in the class $ASM^h \cup ASM^m$ with respect to the properties of existence, relevance and cumulativity, whilst unveiling relations among these properties. As a result of the approach taken in our work, light is shed on the characterization of the SM semantics, as we show that the properties of (lack of) existence and (lack of) cautious monotony are equivalent, which opposes statements on this issue that may be found in the literature. We also characterize the relevance failure of SM semantics in a more clear way than usually stated in the literature.
\end{abstract}


\section{Introduction}
The $SM$ semantics \cite{SM88} is generally accepted by the scientific community working on logic programs semantics as the {\it de facto} standard 2-valued semantics. Nevertheless there are some advantageous properties the SM semantics lacks such as (1) model existence for every \nlp, (2) relevance, and (3) cumulativity \cite{Alex2011}.  Model {\it existence} guarantees that every \nlp has a semantics. This is important to allow arbitrary updates and/or merges involving Knowledge Bases, possibly from different authors or sources \cite{Alex2011}. {\it Relevance} allows for top-down query solving without the need to always compute complete models, but just the sub-models that sustain the answer to a query, though guaranteed extendable to whole ones  \cite{Alex2011}. As for {\it cumulativity}, it allows the programmer to take advantage of tabling techniques \cite{Swift99} for speeding up computations \cite{Alex2011}. Independently of the motivations that underlay the design of a semantics for logic programs, one may ask if it is easy to guarantee some or all of the above properties, or even if it is easy to assess the profile of the resulting semantics in what concerns these properties. In this work we define a family of $2$-valued conservative extensions of the $SM$ semantics, the {\it affix stable model} semantics family, $ASM$. We then take two subclasses, $ASM^h \subset ASM$ and $ASM^m \subset ASM$, and present a number of results that simplify the task of assessing the semantics in $ASM^h \cup ASM^m$ on the properties of existence, relevance and cumulativity. The semantics in these two classes bear resemblance with the already known $SM$ and $MH$ semantics (see section \ref{Families}), and this stands for the motivation to consider them. The following results, obtained in this work, should be emphasized: (1) We present a refined definition of cumulativity for semantics in the class $ASM^h \cup ASM^m$, which turns into an easier job the dismissal of this property by resorting to counter-examples; (2) We divide the sets of rules of \nlps into layers, and use the decomposition of models into that layered structure to define three new (structural) properties, {\it defectivity}, {\it excessiveness} and {\it irregularity}, which allow to state a number of relations between the properties of existence, relevance and cumulativity for semantics of the $ASM^h \cup ASM^m$ class, and at the same time facilitate the assessment of semantics in this class with respect to those properties; (3) As a result of the approach in our work light is shed on the characterization of $SM$ semantics, as we show that the properties of (lack of) existence and (lack of) cautious monotony are equivalent, which opposes statements on this issue that may be found in the literature; we also characterize the relevance failure of $SM$ semantics in a more clear way than usually stated in the literature. It should be stressed that this study is on the properties of a class of 2-valued semantics, under a prospection motivation. The weighing of such semantics rationales under an `intuitive' point of view (or any other equivalently non-objective concept) is beyond the reach of our study. The results presented in this paper are enounced for the universe of finite ground normal logic programs, and are either proved in \cite{abrantesphd}, or immediate consequences of results there contained. \\
\indent The remainder of the paper proceeds as follows. In section \ref{lang_term} we define the language of  \nlps and the terminology to be used in the sequel. In section \ref{Families} the families $ASM$, $ASM^h$ and $ASM^{m}$ are defined. In section \ref{cummulativity} we characterize the property of cumulativity for the families $ASM^h$ and $ASM^{m}$, whilst in section \ref{sec:defexce} the properties of defectivity, excessiveness and irregularity are defined. Some relations among  existence, relevance and cumulativity, which are revealed by means of those properties, are stated. Section \ref{conc} is dedicated to final remarks.

\section{Language and Terminology of Logic Programs} \label{lang_term}
\noindent A {\it \nlp} defined over a language $\mathcal{L}$ is a set of {\it normal} rules, each of the form
\begin{equation}\label{fla:nlprule}
b_0 \< b_1, \cdots, b_m, not\ c_1, \cdots, not\ c_n
\end{equation}
\noindent where $m, n$ are non-negative integers and $b_j, c_k$ are {\it atoms} of  $\mathcal{L}$; $b_i$ and $not\ c_k$ are generically designated {\it literals}, $not\ c_k$ being specifically designated {\it default literal}. The operator `$,$' stands for the conjunctive connective, the operator `$not$' stands for negation by default and the operator `$\<$' stands for a {\it dependency} operator that establishes a dependence of $b_0$ on the conjunction on the right side of `$\<$'. $b_0$ is the {\it head} of the rule and $ b_1, \cdots, b_m, not\ c_1, \cdots, not\ c_n$ is the {\it body} of the rule. A rule is a {\it fact} if $m=n=0$. A literal (or a program) is {\it ground} if it does not contain variables. The set of all ground atoms of a \nlp is called {\it Herbrand base} of $P$, $\mathnormal{H}_P$. A program is {\it finite} if it has a finite number of rules\footnote{In this work, if nothing to the contrary is said, by `logic program', or simply by `program', we mean a finite set of normal ground rules.}. Given a program $P$, program $Q$ is a {\it subprogram} of $P$ if $Q \subseteq P$, where $Q, P$ are envisaged as sets of rules.\\

\noindent For ease of exposition we henceforth use the following abbreviations: $Atoms(E)$, is the set of all {\it atoms} that appear in the ground structure $E$, where $E$ can be a rule, a set of rules or a set of logic expressions; $Body(r)$, is the set of literals in the body of a ground rule $r$; $Facts(E)$, is the set of all facts that appear in the set of rules $E$; $Heads(E)$, is the set of all atoms that appear in the heads of the set of rules $E$; if $E$ is unitary, we may use `$Head$' instead of `$Heads$' . We may compound some of these abbreviations, as for instance $Atoms(Body(r))$ whose meaning is straightforward. Each of the abbreviations may also be taken as the conjunction of the elements contained in the respective sets.\\

\noindent Given a $2$-valued interpretation $I$ of a logic program $P$, we represent by $I^+$ (resp. $I^-$ ) the set of its positive literals (resp. atoms whose default negations are true with respect to $I$). If $I$ is $3$-valued, we additionally represent by $I^u$ the set of undefined atoms with respect to $I$.\\

\noindent The following concepts concern the structure of programs. Let $P$ be a logic program and $r, s$ any two rules of $P$. {\bf Complete rule graph, $\mathbf{CRG(P)}$}\footnote{Adapted from \cite{Alex2011}.}: is the directed graph whose vertices are the rules of $P$. Two vertices representing rules $r$ and $s$ define an arc from $r$ to $s$ iff $Head(r) \subseteq Atoms(Body(s))$. {\bf Rule depending on a rule}$^2$: rule $s$ {\it depends} on rule $r$ iff there is a directed path in $CRG(\mathnormal{P})$ from $r$ to $s$. {\bf Subprogram relevant to an atom}\footnote{Adapted from \cite{journals/fuin/Dix95a}}: a rule $r \in P$ is {\it relevant} to an atom $a \in \mathnormal{H}_P$ iff there is a rule $s$ such that $Head(s)=\{a\}$ and $s$ depends on $r$. The set of all rules of $P$ relevant to $a$ is represented by $Rel_P(a)$, and is named {\it subprogram (of $P$) relevant} to $a$. {\bf Loop}\footnote{Adapted from \cite{CostantiniOlons}}: a set of rules $R$ forms a {\it loop} (or the rules of set $R$ are {\it in loop}) iff, for any two rules $r, s \in R$, $r$ depends on $s$ and $s$ depends on $r$. We say that rule $r \in R$ is in loop through literal $L \in Body(r)$ iff there is a rule $s \in R$ such that $Head(s)=Atoms(L)$. {\bf Rule layering}$^3$: the {\it rule layering} (or just \emph{layering}, for simplicity) of $P$ is the labeling of each rule $r \in P$ with the smallest possible natural number, $layer(r)$, in the following way: for any two rules $r$ and $s$, (1) if rules $r, s$ are in loop, then $layer(r)=layer(s)$; (2) if rule $r$ depends on rule $s$ but rule $s$ does not depend on rule $r$, then $layer(r)>layer(s)$. Every integer number $T$ in the image of the layer function defines a {\it layer} of $P$, meaning the set of rules of $P$ labeled with number $T$ -- we use the expression `layer' to refer both to a set of all rules with that label, and to the label itself. We represent by $P^{\leq T}$ (resp. $P^{> T}$) the set of all rules of $P$ whose layer is less than or equal to (resp. greater than) $T$. {\bf  $\mathbf{T}$-segment of a program}: we say that $P^{\leq T}$ is the {\it T-segment} of $P$ iff  $Atoms(P^{\leq T}) \cap Heads(P^{> T}) = \emptyset$. We may also say `segment $T$' to mean the set of rules corresponding to segment $P^{\leq T}$.\\

\noindent Let $SEM$ be a 2-valued semantics and $SEM(P)$ the set of $SEM$ models of a logic program $P$. Let also the set of atoms $ker_{SEM}(P)=\underset{M \in SEM(P)}{\bigcap} M^+$  be dubbed {\it semantic kernel} of $P$ with respect to $SEM$ (the semantic kernel is not defined if $SEM(P)=\emptyset$). The following properties concern semantics of logic programs. We say that a semantics $SEM$ is: {\bf Existential} iff every \nlp has at least one $SEM$ model; {\bf Cautious monotonic}\footnote{
Adapted from \cite{journals/fuin/Dix95,journals/fuin/Dix95a}.} iff for every \nlp $P$, and for every set $S \subseteq ker_{SEM}(P)$, we have $ker_{SEM}(P) \subseteq ker_{SEM}(P \cup S)$; {\bf Cut}
 iff for every \nlp $P$, and for every set $S \subseteq ker_{SEM}(P)$, we have $ker_{SEM}(P \cup S) \subseteq ker_{SEM}(P)$; {\bf Cumulative}
 iff it is cautious monotonic and cut; {\bf Relevant}
 iff for every \nlp $P$ we have 
\begin{align}\label{flarel}
\forall_{a \in \mathnormal{H}_P} (a \in ker_{SEM}(P) \Leftrightarrow a \in ker_{SEM}(Rel_P(a)))
\end{align}
where $Rel_P(a)$ is the subprogram of $P$ relevant to atom $a$; {\bf Global to local relevant} iff the logical entailment '$\Rightarrow$' stands in formula (\ref{flarel}); {\bf Local to global relevant} iff the logical entailment '$\Leftarrow$' stands in formula (\ref{flarel}). 

\section{Conservative Extensions of the $\mathbf{SM}$ Semantics} \label{Families}

\noindent In this section we define a family of abductive 2-valued semantics\footnote{See \cite{deneckerabduction} for {\it abductive semantics}.}, the {\it affix stable model} family, $ASM$, whose members are conservative extensions of the $SM$ semantics. For that purpose we need the concepts of {\it reduction system} and {\it MH semantics}.

\subsection{Reduction System and $MH$ Semantics} \label{redsystem}
In \cite{brasstransformationbased} the authors propose a set of five operations to reduce a program (i.e., eliminate rules or literals) -- positive reduction, PR, negative reduction, NR, success, S, failure, F and loop detection, L (see \ref{redops} for the definitions of these operations). We represent this set of operations as $\mapsto_{WFS}:=\{PR, NR, S, F, L\}$. By non-deterministically applying this set of operations on a program $P$, we obtain the program $\widehat{P}$, the {\it remainder} of $P$, which is invariant under a further application of any of the five operations. This transformation is {\it terminating} and {\it confluent} \cite{brasstransformationbased}. We denote the transformation of $P$ into $\widehat{P}$ as $P \mapsto_{WFS} \widehat{P}$. We also write $\widehat{P}=remainder_{WFS}(P)$. It is shown in \cite{brasstransformationbased} that $WFM(P)=WFM(\widehat{P})$, where $WFM$ stands for the well-founded model \cite{VanGelder93}. See \ref{remdrcomp} for an example of the computation of the remainder of a program.

\indent One way to obtain conservative extensions of the $SM$ semantics, is to relax some operations of the reduction system $\mapsto_{WFS}$, which yields weaker reduction systems, that is, systems that erase less rules or literals than $\mapsto_{WFS}$. An example of such a semantics is the {\it minimal hypotheses semantics}, $MH$ \cite{Alex2011}, whose reduction system $\mapsto_{MH}$ is obtained from $\mapsto_{WFS}$ by replacing the negative reduction operation, $NR$, by the {\it layered negative reduction} operation, $LNR$, i.e., $\mapsto_{MH}:=\{PR, LNR, S, F, L\}$. $LNR$ is a weaker version of $NR$ that instead of eliminating any rule $r$ containing say $not\ b$ in the body, in the presence of the fact $b$, as $NR$ does, only eliminates rule $r$ if this rule is not in loop through literal $not\ b$. We write $P \mapsto_{MH} \mathring{P}$, where $\mathring{P}$ is the {\it layered remainder} of $P$. We also write $\mathring{P}=remainder_{MH}(P)$. See \ref{mhmodelscomp} for an example of the computation of the layered remainder of a program.

\subsection{$\mathbf{ASM, ASM^h}$ and $\mathbf{ASM^m}$ Families}

We define {\it affix stable interpretation} and then use this concept to put forward the definition of $ASM$ family. 

\mydef{Affix Stable Interpretation.}{def:stbint}{Let P be a \nlp, $SEM$ a 2-valued semantics with a corresponding reduction system $\mapsto_{SEM}$, and $X \subseteq Atoms(remainder_{SEM}(P))$. We say that $I$ is an {\it affix stable interpretation} of P with respect to set X and semantics $SEM$ (or simply a {\it $SEM$ stable interpretation with affix X}) iff $I=WFM(P \cup X)$ and $WFM^u(P \cup X)=\emptyset, $\footnote{Notice that $WFM^u(P \cup X)$ is the set of undefined atoms in the model $WFM(P \cup X)$.} that is, $I$ is the only stable model of the program $P \cup X$. We name $X$ an {\it affix} (or {\it hypotheses set}) of interpretation $I$. We also name {\it assumable hypotheses} set of program $P$, $Hyps(P)$, the union of all possible affixes that may be considered to define the stable interpretations (we have $Hyps(P) \subseteq Atoms(remainder_{SEM}(P))$).}


\mydef{Affix Stable Model Semantics Family, $\mathbf{ASM}$.}{def:afxsm}{A 2-valued semantics $SEM$, with a corresponding reduction system $\mapsto_{SEM}$, belongs to the {\it affix stable model} semantics family, $ASM$, iff, given any \nlp $P$, $SEM(P)$ contains all the $SM$ models of $P$, in case they exist, plus a subset (possibly empty) of the affix stable interpretations of $P$ chosen by resorting to specifically enounced criteria.}

\indent Both semantics $SM$ and $MH$ belong to the $ASM$ family. The two non-disjoint subfamilies of $ASM$ next defined, $ASM^h$ and $ASM^m$, will be the classes whose formal properties we study in the sequel. 

\mydef{$\mathbf{ASM^h}$ and $\mathbf{ASM^m}$ Families.}{def:ASMhm}{A semantics $SEM \in ASM$ belongs to the $ASM^h$  or $ASM^m$ families iff, for any \nlp $P$, the models are computed as follows: 
\begin{enumerate} 
\item For both $ASM^h$  and $ASM^m$ the set of {\it assumable hypotheses}, $Hyps(P)$,  is contained in the set of atoms that appear default negated in $remainder_{SEM}(P)$\footnote{The purpose of computing the remainder of a program, is to obtain the assumable hypotheses set of the program.};
\item For semantics in the class $ASM^h$, the affixes of the models of $P$ are either those non empty minimal with respect to set inclusion, if $Hyps(P) \neq \emptyset$, or else the empty set if $Hyps(P)=\emptyset$. For semantics in the class $ASM^m$, the  models in $SEM(P)$ are always minimal models.
\end{enumerate}
}

We now refer some examples of $ASM^h$ and $ASM^m$ members, whose definitions can be found in \ref{asmsemantics}. Besides $SM$, $MH$ and others, the following are $ASM^h$ family members, referred to subsequently:\footnote{The first three semantics were suggested by Alexandre Pinto.} $\mathbf{MH^{LS}}$, $\mathbf{MH^{Loop}}$, $\mathbf{MH^{Sustainable}}$, $\mathbf{MH^{Sustainable}_{min}}$, $\mathbf{MH^{Regular}}$. Besides $SM$ and others, the following are $ASM^m$ family members, referred to subsequently: $\mathbf{Navy}$, $\mathbf{Blue}$, $\mathbf{Cyan}$, $\mathbf{Green}$.

\section{Characterization of Cumulativity for the $\mathbf{ASM^h \cup ASM^m}$ Class}	\label{cummulativity}
\noindent In this section we lay down a characterization of cumulativity for semantics $SEM$ of the $ASM^h \cup ASM^m$ class, via the following theorem. 

\mytheo{}{teo:equalcum}{Let $SEM$ be a semantics of the $ASM^h \cup ASM^m$ class. For every program $P$ and for every subset $S \subseteq ker_{SEM}(P)$, the following results stand: (1) $SEM$ is cautious monotonic iff $SEM(P \cup S) \subseteq SEM(P)$; (2) $SEM$ is cut iff $SEM(P) \subseteq SEM(P \cup S)$; (3) $SEM$ is cumulative iff $SEM(P)=SEM(P \cup S)$ -- this is a consequence of statements (1) and (2).\footnote{Notice that $SEM(P)$ represents the set of all $SEM$ models of $P$.}}

\indent The three items of this theorem correspond to refinements of the classical definitions of cautious monotony, cut and cumulativity  (see section \ref{lang_term}). The new definitions establish the properties by means of relations among sets of models, as opposed to the relations among sets of atoms that characterize the classical definitions.\\

\noindent The results stated in this theorem are advantageous to spot cumulativity failure in semantics of the $ASM^h \cup ASM^m$ class by means of counter-examples (logic programs), when compared with common procedures (e.g., \cite{journals/fuin/Dix95,journals/fuin/Dix95a}). The reason is that common procedures always need the counter-examples to fail cumulativity\footnote{The general procedure to spot the failure of cumulativity by resorting to counter-examples is as follows: compute all the $SEM$ models of a program $P$; add to $P$ subsets $ S \subseteq ker_{SEM}(P)$, and compute all the models of the resulting programs $P \cup S$, drawing a conclusion about cumulativity failure only in cases where $ker_{SEM}(P) \neq ker_{SEM}(P \cup S)$.}, whilst the results of theorem \ref{teo:equalcum} allow us to spot failure of cumulativity even in some cases where the counter-examples used do not show any failure of this property. To make this point clear see the examples in  \ref{cumfailure} and  \ref{cmcutfailure}.\\

\noindent It should be stressed that there are $2$-valued cumulative semantics to which $SEM(P) \neq SEM(P \cup S)$ for some \nlp $P$ and some $S \subseteq ker_{SEM}(P)$ (for an example, see the definition of the 2-valued semantics {\it Picky} in \ref{picky}). Theorem \ref{teo:equalcum} states this is not the case if $SEM \in ASM^h \cup ASM^m$.

\section{Defectivity, Excessiveness and Irregularity}\label{sec:defexce}

\noindent \noindent Theorem \ref{teo:equalcum} application for dismissing the cumulativity property by means of counter-examples, demands computing the set of models $SEM(P)$ of a program $P$, the set $ker_{SEM}(P)$, and after this it needs the computation of the sets of models $SEM(P \cup S)$, $S \in ker_{SEM}(P)$, to look for a case that eventually makes $SEM(P)=SEM(P \cup S)$ false. In this section three structural properties are defined, {\it  defectivity, excessiveness} and {\it irregularity}, that will turn the dismissal of existence, relevance or cumulativity spottable by means of one model only. It will be shown that for semantics of the $ASM^h \cup ASM^m$ class, defectivity is equivalent to the failure of existence and to the failure of global to local relevance, and also entails the failure of cautious monotony, whilst excessiveness entails the failure of cut, and irregularity is equivalent to the failure of local to global relevance. 

\subsection{Defectivity}
\noindent The rationale for the concept of defective semantics is the following: if a segment $P^{\leq T}$ has a  $SEM$ model $M$ that is not contained in any whole $SEM$ model of $P$, then we say the semantics $SEM$ is {\it defective}, in the sense that it `does not use' all the models of segment $T$ in order to get whole models of $P$. 

\mydef{Defective semantics.}{def:defective}{A 2-valued semantics $SEM$ is called {\it defective} iff there is a \nlp $P$, $SEM(P) \neq \emptyset$, a segment $P^{\leq T}$ of $P$, and a $SEM$ model $M$ of the segment $P^{\leq T}$, such that $SEM(P^{> T} \cup M^+) =\emptyset$. We also say that $SEM$ is {\it defective} with respect to segment $T$ of program $P$, and that $M$ is a {\it defective} model of $P$ with respect to segment $T$ and semantics $SEM$.}

\myex{}{ex:smnotexist}{Program $P=\{a\<not\ b,  \:b\<not\ a,  \:c\<a,  \:c\<not\ c\}$ may be used to show that the $SM$ semantics is defective. In fact, the only $SM$ model of $P$ is $N=\{a, not\ b, c\}$ with affix $\{a\}$. Meanwhile, $P^{\leq 1}=\{a \< not\ b, \:b\< not\ a\}$ is a segment that has the stable model $M=\{not\ a, b\}$, and we have $SM(P^{> 1} \cup \{b\})=\emptyset$.}

\indent The next theorem shows how conclusions about existence, relevance and cumulativity may be immediately taken in the case of a defective semantics.

\mytheo{}{teo:equivdefect}{The following relations are valid for any semantics of the $ASM^h \cup ASM^m$ class: 
\begin{flalign*}
1.&\; \text{Defectivity} \Leftrightarrow \neg \text{Existence}  \Leftrightarrow \neg \text{Global to Local Relevance};\\2.&\;  \text{Defectivity} \Rightarrow \neg \text{Cautious Monotony}.
\end{flalign*}}

\indent The reader should notice the importance of this theorem: not only defectivity is enough to dismiss existence, relevance and cumulativity, as also these properties appear strongly related for semantics of the class $ASM^h \cup ASM^m$: if existence fails then relevance also fails (through global to local relevance failure); if existence fails then cumulativity also fails (through cautious monotony failure); if relevance fails (through global to local relevance failure), then cumulativity also fails (through cautious monotony failure). Definition \ref{def:defective} above shows the structural nature of defectivity, which allows the verification of the property by wisely constructing a program that satisfies it. This may turn easier the assessment of existence, relevance and cumulativity, when compared to dealing with this issue on the basis of abstract proofs. Even more, the relation between existence and defectivity stated in theorem \ref{teo:equivdefect}, allows the failure of the existence property to be detected by resorting to counter-examples, even in some cases where the program used as counter-example has models. E.g., program $P$ in \ref{cumfailure} can be used to detect the failure of existence for $SM$ semantics, in spite of the existence of stable models for program $P$, since it reveals the defectivity of  $SM$.\footnote{It should be pointed out  that there are 2-valued semantics for which the equivalence $defectivity \Leftrightarrow \neg existence$ fails, e.g., $M^{Supp}_P$ \cite{abw87} which is not defective in spite of failing the existence property -- it is the case that $M^{Supp}_P$ is not a $ASM$ semantics, since it does not conservatively extend the $SM$ semantics.}\\

\noindent The results stated in theorem \ref{teo:equivdefect} also shed some light on the characterization of $SM$ semantics with respect to the properties of existence and cumulativity. In \cite{journals/fuin/Dix95a}, section $5.6$, the author says that the $SM$ is not cumulative and that this fact does not depend on the non existence of stable models (i.e., the author states that lack of cumulativity is not a consequence of lack of existence). Meanwhile theorem \ref{teo:equivdefect} above shows that $SM$ is non-existential due to being defective, which in turn makes it not cautious monotonic and thus not cumulative. Thus the failure of cumulativity for the $SM$ semantics case is indeed a consequence of the failure of existence for this semantics. Moreover, with respect to the $SM$ semantics a stronger result relating existence and cautious monotony may be enounced: these two properties show up equivalence in the sense stated in proposition \ref {prop:smexistcm} below. To the best of our knowledge, this connection between these two properties had not yet been stated. 

\myprop{}{prop:smexistcm}{For the $SM$ semantics the following result stands: there is a program $P$ that shows existence failure iff  there is a program $P^*$ that shows cautious monotony failure.}

 
\subsection{Excessiveness and Irregularity}
\noindent The rationale of the concept of {\it excessive}  semantics is the following: if a \nlp $P$ has a model $N$ and a layer $P^{\leq T}$ such that for every model $M_* \in SEM(P^{\leq T})$ it is the case that $N \notin SEM(P^{> T} \cup M^+_*)$, then we say that model $N$ (and thus the semantics) is {\it excessive}, in the sense that it `goes beyond' the semantics of  the segment $P^{\leq T}$ by not being a `consequence' of it. 

\mydef{Excessive semantics.}{def:excessive}{A 2-valued semantics $SEM$ is called {\it excessive} iff there is a logic program $P$, a segment $P^{\leq T}$, a model $M \in SEM(P^{\leq T})$ and a model $N \in SEM(P)$ such that:
\begin{enumerate}
\item \label{def:excpoint1}$M^+=N^+_{\leq T}$, where $N^+_{\leq T}=N^+ \cap Heads(P^{\leq T})$;
\item For every model $M_* \in SEM(P^{\leq T})$ it is the case that $N \notin SEM(P^{> T} \cup M^+_*)$;
\item There is at last a $SEM$ model $N^*$ of $P$, such that $N^* \in SEM(P^{>T} \cup M^+)$.
\end{enumerate}
\noindent We also say that $SEM$ is {\it excessive} with respect to segment $T$ of program $P$, and that $N$ is an {\it excessive model} of $P$ with respect to segment $T$ and semantics $SEM$.}

\indent In the excessiveness example in \ref{excirreg} it is shown that the semantics $MH, MH^{LS}, MH^{Loop},$ $Navy, Green$ are excessive.\\

\noindent The rationale of the concept of {\it irregularity} is as follows: given a certain whole model $N \in SEM(P)$, if the set $N^+ \cap Heads(P^{\leq T})$ is not a model of a segment $P^{\leq T}$, then we say that $SEM$ is {\it irregular}, since $N$ `is not a consequence' of the semantics of segment $T$.

\mydef{Irregular semantics.}{def:irregular}{A 2-valued semantics $SEM$ is called {\it irregular} iff there is a logic program $P$, a segment $P^{\leq T}$ and a $SEM$ model $N$ of $P$, such that for  no model $M$ of $P^{\leq T}$ do we have $N^+_{\leq T}=M^+$, where $N^+_{\leq T}=N^+ \cap Heads(P^{\leq T})$. We also say that $SEM$ is {\it irregular} with respect to segment $T$ of program $P$, and that $N$ is an {\it irregular} model of $P$ with respect to segment $T$ and semantics $SEM$. A model that is not irregular is called {\it regular}, and a semantics that produces only regular models is called {\it regular}.\footnote{In comparing excessiveness and irregularity, notice that a whole model can be excessive whilst containing models for all the segments of the program (i.e., be a regular model) - see the excessiveness example in \ref{excirreg}.}}

\indent The concepts of excessiveness and irregularity exhibit independence for semantics of the $ASM^h \cup ASM^m$ class, meaning there is a semantics in this class for any of the four possible cases of validity or failure of excessiveness and irregularity. As a matter of fact, it can be shown \cite{abrantesphd} that $Blue$ is irregular whilst not excessive (i.e., $irregularity \nRightarrow excessiveness$); it is also the case that $MH^{Regular}$ is excessive but not irregular  (i.e., $excessiveness \nRightarrow irregularity$). Also $MH$ is excessive and irregular, and $Cyan$ is not excessive and is not irregular. \\

\noindent The following result states relations between excessiveness and cut, and between irregularity and relevance.

\mytheo{}{teo:cutexcelgr}{ The following relations stand  for  any semantics of the $ASM^h \cup ASM^m$ class:
\begin{align*}
1.&\;  \text{Excessiveness } \Rightarrow \neg  \text{Cut};\\
2.&\; \text{Irregularity} \Leftrightarrow \neg \text{Local to Global Relevance}.
\end{align*}
}

\indent As excessiveness and irregularity are structural properties, being thus detectable by construction of adequate programs, they facilitate, via this theorem, the dismissal of cut and relevance. For instance, this result together with the excessiveness example in \ref{excirreg}, shows that semantics $MH$, $MH^{LS}$, $MH^{Loop}$, $Navy$ and $Green$ are excessive, and thus not cut. Also, this result together with the irregularity example in \ref{excirreg}, shows that semantics $MH$, $MH^{LS}$ and $MH^{Loop}$, $Green$, $Navy$ and $Blue$ are irregular, and thus not relevant. As was the case for the relation between the properties of existence and cumulativity for the $SM$ semantics, our work sheds also some light on the $SM$ semantics relevance failure, through the following results. 

\myprop{}{prop:smnotexc}{Let $P$ be a \nlp and $M \in SM(P)$. Then $M$ is neither excessive nor irregular.}
\mycor{}{cor:smvacrel}{$SM$ is (vacuously) local to global relevant.}

\indent Notice that this corollary, together with the example in \ref{cumfailure} and theorem \ref{teo:equivdefect},  let clear the cause for $SM$ semantics relevance failure: $SM$ fails relevance because it fails global to local relevance. This is a more precise characterization than just saying that $SM$ is not relevant, as usually stated in literature (e.g., \cite{journals/fuin/Dix95a}). \\

\noindent If we consider the five formal properties of existence ($\exists$), global to local relevance ($gl$), local to global relevance ($lg$), cautious monotony ($cm$) and cut ($cut$), the validity or failure of each of these properties allow, in the general case, the existence of $2^5=32$ types of semantics. Meanwhile, the study we present in this work shows that only $12$ such types of semantics may exist in the $ASM^h \cup ASM^m$ class. They are represented in table \ref{tab:12types} in \ref{twelvetypes}.


\section{Final Remarks}	\label{conc}
\noindent In this paper we considered the characterization of 2-valued conservative extensions of the $SM$ semantics on the properties of existence, relevance and cumulativity. This theoretical endeavor is reasonable under a point of view of prospectively assessing the behavior of such types of semantics with respect to a set of properties that are desirable, both under a computational (relevance and cumulativity) and a semantical (existence) standpoint. For that purpose we focused our study on two subsets of the here defined $ASM$ class of 2-valued conservative extensions of the $SM$ semantics, the non-disjoint classes $ASM^h$ and $ASM^m$, whose elements maintain a degree of resemblance with already known 2-valued semantics, such as the $SM$ and the $MH$ semantics. As a result of this study, refined definitions of cautious monotony, cut and cumulativity were set. This new definitions turn into an easier job the dismissal of the properties of existence, relevance and cumulativity, as shown in section \ref{cummulativity}. This study also reveals relations among these properties, unveiled by theorems \ref{teo:equivdefect} and \ref{teo:cutexcelgr}, that allow to draw conclusions about some of them on basis of held knowledge about others. This last point builds on top of the new structural properties of defectivity, excessiveness and irregularity, which provide an analytical shortcut to assess existence, relevance and cumulativity. The approach taken in this work (characterizing families of semantics, instead of individual semantics), revealed itself advantageous also in clarifying the profile of the well known and studied $SM$ semantics, via the results stated in proposition \ref{prop:smexistcm} and corollary \ref{cor:smvacrel}. Our work also states a maximum of 12 types of semantics in the class $ASM^h \cup ASM^m$, with respect to the satisfaction/failure of the properties of existence ($\exists$), global to local relevance ($gl$), local to global relevance ($lg$), cautious monotony ($cm$) and cut ($cut$).\\

\noindent Finally, the structural approach put forward in this paper has the potential of being used with semantics other than 2-valued ones, and with other strong and weak properties besides existence, relevance or cumulativity.\footnote{The terms {\it strong} and {\it weak} applied to formal properties, are here adopted after \cite{journals/fuin/Dix95,journals/fuin/Dix95a}.}

\section*{Acknowledgments}
We thank Alexandre Pinto for some important debates on conservative extensions of the $SM$ semantics. The work on this paper has been partially supported by Funda\c{c}\~ao para a Ci\^encia e Tecnologia and Instituto Polit\'ecnico de Bragan\c{c}a grant PROTEC : SFHR/49747/2009.

\bibliographystyle{acmtrans}
\bibliography{ICLP14_CameraReady}


\newpage
\appendix

\section{Reduction Operations} \label{redops} 
 In the definitions below, $P_{1}$ and $P_{2}$ are two ground logic programs.
\begin{enumerate}
\item {\bf Positive reduction, PR.} Program $P_{2}$ results from $P_{1}$ by \emph{positive reduction} iff there is a rule $r \in P_{1}$ and a default literal $not\ b \in Body(r)$ such that $b\notin Heads(P_{1})$, and $P_{2}=(P_{1}\setminus\{r\})\cup\{Head(r)\<(Body(r) \setminus \{not\ b\})\}$.
\item{\bf Negative reduction, NR.} Program $P_{2}$ results from $P_{1}$ by \emph{negative reduction} iff there is a rule $r \in P_{1}$ and a default literal $not\ b \in Body(r)$ such that $b \in Facts(P_{1})$, and $P_{2}=P_{1}\setminus\{r\}$.
\item{\bf Success, S.} Program $P_{2}$ results from $P_{1}$ by \emph{success} iff there is a rule $r \in P_{1}$ and a fact $b \in Facts(P_{1})$ such that $b\in Body(r)$, and $P_{2}=(P_{1}\setminus\{r\})\cup\{Head(r)\<(Body(r)\setminus\{b\})\}$.
\item{\bf Failure, F.} Program $P_{2}$ results from $P_{1}$ by \emph{failure} iff there is a rule $r \in P_{1}$ and a positive literal $b\in Body(r)$ such that $b \notin Heads(P_{1})$, and $P_{2}=P_{1}\setminus\{r\}$.
\item{\bf Loop Detection, L.} Program $P_{2}$ results from $P_{1}$ by \emph{loop detection} iff there is a set $\mathcal{A}$ of ground atoms such that:
\begin{enumerate}
\item For each rule $r \in P_{1}$, if $Head(r) \in \mathcal{A}$, then $Body(r) \cap \mathcal{A}\neq\emptyset$;
\item $P_{2}:=\{r \in P_{1}|Body(r) \cap \mathcal{A}=\emptyset\}$.
\end{enumerate}	
\end{enumerate}

\section{Remainder Computation Example} \label{remdrcomp} 
Let $P$ be the set of all rules below. The remainder $\widehat{P}$ is the non shadowed part of the program. The labels (i)--(v) indicate the operations used in the corresponding reductions: (i) PR, (ii) NR, (iii) S, (iv) F, (v) L.
{\small $$\{a \< \tikz[baseline] { \node[fill=gold!20, anchor=base, rounded corners =2pt] (d2) {$not\ f$};} \textnormal{(i)}, \;\;		\tikz[baseline] { \node[fill=gold!20, anchor=base, rounded corners =2pt] (d2) {$e$};} \tikz[baseline] { \node[fill=gold!20, anchor=base, rounded corners =2pt] (d2) {$\< d$};}\textnormal{(v)}, \;\;
a \< \tikz[baseline] { \node[fill=gold!20, anchor=base, rounded corners =2pt] (d2) {$not\ b$};}\textnormal{(i)}, \;\;  \tikz[baseline] { \node[fill=gold!20, anchor=base, rounded corners =2pt] (d2) {$d$};} \tikz[baseline] { \node[fill=gold!20, anchor=base, rounded corners =2pt] (d2) {$\< e$};}\textnormal{(v)}, \;\;
\tikz[baseline] { \node[fill=gold!20, anchor=base, rounded corners =2pt] (d2) {$b$};} \tikz[baseline] { \node[fill=gold!20, anchor=base, rounded corners =2pt] (d2) {$\< not\ a$};}\textnormal{(ii)}, \;\;
c \< \tikz[baseline] { \node[fill=gold!20, anchor=base, rounded corners =2pt] (d2) {$a$};}\textnormal{(iii)}, \;\;
\tikz[baseline] { \node[fill=gold!20, anchor=base, rounded corners =2pt] (d2) {$d$};} \tikz[baseline] { \node[fill=gold!20, anchor=base, rounded corners =2pt] (d2) {$\< f$};}\textnormal{(iv)}\}$$}


\section{Minimal Hypotheses Models Computation} \label{mhmodelscomp} 
Let $P$ be the set of rules below, which is equal to the program in \ref{remdrcomp}. The layered remainder $\mathring{P}$ is the non-shadowed part of the program.
\begin{align*}
a &\< \tikz[baseline] { \node[fill=gold!20, anchor=base, rounded corners =2pt] (d2) {$not\ f$};}& d&\<f\\ 
a&\< not\ b & e &\< d\\
b&\< not\ a&d&\<e&\\
c&\<\tikz[baseline] { \node[fill=gold!20, anchor=base, rounded corners =2pt] (d2) {$a$};}&&
\end{align*}
Notice that rule $b \< not\ a$ is no longer eliminated by the fact $a$, since this rule and rule $a\< not\ b$ are in loop, and in the case of rule $b \< not\ a$ the loop is through the literal $not\ a$.\\
\indent The $MH$ models of a program $P$ are computed as follows: (1) Take as assumable hypotheses set, $Hyps(P)$, the set of all atoms that appear default negated in $\mathring{P}$; in the case of the previous program we have $Hyps(P)=\{a, b\}$; (2) Form all programs $P \cup H$, for all possible subsets $H \subseteq Hyps$, $H \neq \emptyset$ (if $Hyps=\emptyset$, then $H=\emptyset$ is the only set to consider); take all the interpretations for which $WFM(P \cup H)$ is a total model (meaning a model that has no undefined literals); $H$ is the {\it hypotheses set} of the interpretation $WFM(P \cup H)$; (3) Take all the interpretations obtained in the previous point, and chose as $MH$ models the ones that have minimal $H$ sets with respect to set inclusion. The $MH$ models of program $P$ in the example above, and the corresponding hypotheses sets, are
\begin{align*}
M_1&=\{a, not\ b, c, not\ d, not\ e, not\ f\}& \quad H&=\{a\}\\
M_2&=\{a, b, c, not\ d, not\ e, not\ f\}&\quad H&=\{b\}.
\end{align*}
Notice that $M_1$ is the only $SM$ model of $P$. The $MH$ reduction system keeps some loops intact, which are used as choice devices for generating $MH$ models, allowing us to have $MH(P) \supseteq SM(P)$. The sets $H$ considered may be taken as abductive explanations \cite{deneckerabduction} for the corresponding models. 

\section{Definitions of some Elements of  $ASM^h$ and $ASM^m$ Families} \label{asmsemantics} 

 Besides $SM$ and others, the following are $ASM^h$ family members.\\
\noindent $\mathbf{MH^{LS}}$: the reduction system is obtained by replacing the success operation in $\mapsto_{MH}$ by the {\it layered success} operation;\footnote{{\it Layered success} is an operation proposed by Alexandre Pinto. It weakens the operation of success by allowing it to be performed only in the cases where the rule $r$, whose body contains the positive literal $b$ to be erased, is not involved in a loop through literal $b$.} $MH^{LS}$ models are computed as in the $MH$ case.\\
$\mathbf{MH^{Loop}}$: the reduction system is  $\mapsto_{MH}$; the assumable hypotheses set of a program $P$, $Hyps(P)$, is formed by the atoms that appear default negated in literals involved in loops in the  layered remainder $\mathring{P}$; $MH^{Loop}$ models are computed as in the $MH$ case.\\
$\mathbf{MH^{Sustainable}}$: the reduction system is  $\mapsto_{MH}$; $MH^{Sustainable}$ models are computed as in the $MH$ case with the following additional condition: if $H$ is a set of hypotheses of a $MH^{Sustainable}$ model $M$ of $P$, then $$\forall_{h \in H} \;[(H\setminus \{h\}) \neq \emptyset \Rightarrow h \in WFM^u(P \cup (H \setminus\{h\}))],$$ that is, no single hypothesis may be defined in the well-founded model if we join to $P$ all the other remaining hypotheses.\\
$\mathbf{MH^{Sustainable}_{min}}$: the reduction system is  $\mapsto_{MH}$; $MH^{Sustainable}_{min}(P)$ retrieves the minimal models contained in $MH^{Sustainable}(P)$ for any \nlp $P$. $MH^{Sustainable}_{min}$ also belongs to the $ASM^m$ family, due to the minimality of its models.\\
$\mathbf{MH^{Regular}}$: the reduction system is  $\mapsto_{MH}$; retrieves the same models as $MH$, except for the irregular ones (cf. Definition \ref{def:irregular}).\\

\noindent Besides $SM$, $MH^{Sustainable}_{min}$ (defined above in this appendix) and others, the following are $ASM^m$ family members.\\
\noindent $\mathbf{Navy}$: the reduction system is $\mapsto_{WFS}$. Given a \nlp $P$, $Navy(P)$ contains all the minimal models of $\widehat{P}$.\footnote{See definition of $\widehat{P}$ in subsection \ref{redsystem}.}\\
$\mathbf{Blue}$: the reduction system is $\mapsto_{WFS}$. Given a \nlp $P$, $Blue(P)$ contains all the models in $Navy(P \cup K)$ where $K$ is obtained after terminating the following algorithm:\footnote{This algorithm is presented in \cite{journals/fuin/Dix95}.} \\(a) Compute $K=kernel_{Navy}(\widehat{P})$; \\(b) Compute $K^{\prime}=kernel_{Navy}(\widehat{P \cup K})$; \\(c) 	If $K \neq K^{\prime}$, then let $P$ be the new designation of program $P \cup K^{\prime}$; go to step (a). \\Repeat steps (a) -- (c) until $K \neq K^{\prime}$ comes false in (c).\\
$\mathbf{Cyan}$: the reduction system is $\mapsto_{WFS}$. Given a \nlp $P$, compute $Cyan(P)$ through the steps of $Blue$ computation, but taking only the {\it regular} models (cf. Definition \ref{def:irregular}) to compute the semantic kernel at steps (a) and (b).\\
$\mathbf{Green}$: the reduction system is $\mapsto_{WFS}$. Given a \nlp $P$, $Green(P)$ contains all the minimal models of $\widehat{P}$ that have the smallest (with respect to set inclusion) subsets of classically unsupported atoms.\footnote{Given a logic program $P$, a model $M$ of $P$ and an atom $b \in M$, we say that $b$ is classically unsupported by $M$ iff there is no rule $r \in P$ such that $Head(r)=\{b\}$ and all literals in $Body(r)$ are true with respect to $M$.}

\section{Example of Cumulativity Failure Detection} \label{cumfailure} 
The following $1$-layer program $P$ is a counter-example for showing, using theorem \ref{teo:equalcum}, that $SM$ semantics is not cumulative, due to being not cautious monotonic (program $P$ does not allow us to spot the failure of any of these properties by means of the usual definitions of cumulativity and cautious monotony  presented in section \ref{lang_term}) .
\begin{align*}
										a&\<not\ b, not\ s												&d&\<b								&d&\<a	\\
										b&\< not\ a, not\ c											&d&\<not\ d 				&c&\<k	\\
										c&\<not\ b, not\ k											&k&\<a,d						&s&\<not\ a, d
\end{align*}
In fact, the $SM$ models of $P$ are $\{a,d, c, k\}$ and $\{b,d, s\}$, and thus $ker_{SM}(P)=\{d\}$. Now $P \cup \{d\}$ has the stable models $\{a,d, c, k\}$, $\{b,d,s\}$ and $\{c,d, s\}$, and thus $ker_{SM}(P)=ker_{SM}(P \cup \{d\})=\{d\}$. Hence no negative conclusion can be afforded about cumulativity, by means of the usual definition of this property. Meanwhile, by using the statement (3) of theorem \ref{teo:equalcum} it is straightforward to conclude that $SM$ semantics does not enjoy the property of cumulativity, because $SM(P) \neq SM(P \cup \{d\})$. Moreover, statement (1) of the theorem tells us, via this example, that $SM$ semantics is not cautious monotonic because $SM(P \cup \{d\}) \nsubseteq SM(P)$.

\section{Proof of Cautious Monotony and Cut Failure} \label{cmcutfailure} 

\noindent The following $1$-layer program $P=\mathring{P}$ is a counter-example for showing, using theorem \ref{teo:equalcum}, that none of the semantics $MH$, $MH^{LS}$, $MH^{Loop}$, $MH^{Sustainable}$ and $MH^{Regular}$ is either cautious monotonic or cut (program $P$ does not allow us to spot the failure of any of these properties by means of the usual definitions of cautious monotony and cut presented in section \ref{lang_term}) .
\begin{align*}
										u&\<b									&a &\<not\ b					\\
										u&\<c									&b &\<not\ c 				\\
										t&\<a										&c &\<h, u							\\
										t&\<h										&h &\<not\ h, not\ t	
\end{align*}
\noindent Let $SEM$ represent any of the above semantics. The minimal hypotheses models are the same with respect to any of the four semantics (models are represented considering only positive literals): $\{c, u, a, t\}$  with affix  $\{c\}$; $\{b, h, u, c, t\}$ with affix $\{b, h\}$; $\{t, b, u\}$ with affix $\{t\}$. Thus $ker_{SEM}(P)=\{t, u\}$. Now it is the case that the remainder of $P \cup \{u\}$ is the same for any of these semantics: 
\begin{align*}
										u&\<b									&a &\<not\ b					\\
										u&\<c									&b &\<not\ c 				\\
										t&\<a										&c &\<h								\\
										t&\<h										&h &\<not\ h, not\ t	 &u&\<
\end{align*}
\noindent (as a matter of fact, the remainder for the $MH^{LS}$ has the rule $c \< h, u$ instead of $c \< h$; but this does not change the sequel of this reasoning). The minimal hypotheses models of $P \cup \{u\}$ are the same with respect to any of the four semantics (models are represented considering only positive literals): $\{c, u, a, t\}$  with affix  $\{c\}$; $\{h, u, c, t, a\}$ with affix $\{h\}$; $\{t, b, u\}$ with affix $\{t\}$. Thus $ker_{SEM}(P \cup \{u\})=\{t, u\}=ker_{SEM}(P)$, and no conclusions about cumulativity can be drawn by means of the usual general procedures. Meanwhile, $M=\{h, u, c, t, a\}$, with affix $\{h\}$, is a minimal affix model of $P \cup \{u\}$ but is not a minimal affix model of $P$, which by point (1) of theorem \ref{teo:equalcum} renders any of these semantics not cautious monotonic. Also $N=\{b, h, u, c, t\}$, with affix $\{b, h\}$, is a minimal affix model of $P$, but not a minimal affix model of $P \cup \{u\}$, which by point (2) of theorem \ref{teo:equalcum} renders any of these semantics as not cut.

\section{Picky, a Special 2-valued Cumulative Semantics} \label{picky} 
The semantics $Picky$ is defined as follows: for any \nlp $P$ (1) if  $SM(P) = \emptyset$, then $Picky(P)=\emptyset$; (2) if $SM(P) \neq \emptyset$, then (2a) $Picky(P)=SM(P)$ iff $ker_{SM}(P)=ker_{SM}(P \cup S)$, for every $S \subseteq ker_{SM}(P)$; (2b) otherwise $Picky(P)=\emptyset$. This semantics is cumulative, by definition, but it is not always the case that $Picky(P) = Picky(P \cup S)$, $S \subseteq ker_{SM}(P)$ : for program $P$ of the example in \ref{cumfailure}, we have $Picky(P)=\{\{a,d, c, k\}, \{b,d, s\}\}$ and $Picky(P \cup \{d\})=\{\{a,d, c, k\}, \{b,d,s\}, \{c,d, s\}\}$, which means, by theorem \ref{teo:equalcum}, that $Picky$ is not cumulative. Notice that $Picky$ is not a $ASM$ semantics, because it does not conservatively extend the $SM$ semantics: for program $P$ in the referred example, we have $SM(P) \neq \emptyset$ and $Picky(P)= \emptyset$.

\section{Excessiveness and Irregularity} \label{excirreg} 

\noindent {\bf Excessiveness.} The following program $P$ shows that semantics $MH$, $MH^{LS}$, $MH^{Loop}$, $Navy$ and $Green$ are excessive (the dashed lines divide the program into layers; top layer is layer 1, bottom layer is layer 4), 
\begin{align*}
a &\< not\ b\\
b &\< not\ a\\
-&----1\\
u&\< a\\
u&\< b\\
-&----2\\
p&\< not\ p, not\ u\\
-&----3\\
q&\< not\ q, not\ p\\
-&----4.
\end{align*}
\noindent Let $SEM$ represent any of these semantics. It is the case that $N=\{a, u, p, not\ b, not\ q\}$ with affix $\{a,p\}$, is a model of $P$ under any of the referred semantics, and for no $SEM$ model $M_* \in SEM(P^{\leq 2})$, where $SEM(P^{\leq 2})=\{\{a, not\ b, u\},\{not\ a, b, u\}\}$,  do we have $N \in SEM(P^{> 2} \cup M^+_*)$, because atom $u \in M^+_*$ eliminates the rule in layer $3$ via layered negative reduction operation (which has here the same effect as negative reduction operation), and thus $p$ belongs to no model in $SEM(P^{>2} \cup M^+_*)$.\\\\

\noindent {\bf Irregularity.} Program $P$ below shows that the semantics $MH$, $MH^{LS}$ and $MH^{Loop}$, $Green$, $Navy$ and $Blue$ are all irregular.
\begin{align*}
a &\< not\ b\\
b &\< not\ a\\
-&----1\\
p&\< not\ p, not\ a\\
q&\< not\ q, not\ b
\end{align*}
\newline In fact, all these semantics admit the model $N=\{a, b, not\ p, not\ q\}$. The models of segment $P^{\leq 1}$ are $\{a, not\ b\}$ and $\{b, not\ a\}$, none of whose positive sets of atoms equals $N_{\leq T}^+=\{a, b\}$. As $Blue$ is not excessive, this example shows $irregularity \nRightarrow excessiveness$.

\section{The 12 possible types of $ASM^h$ and $ASM^m$ semantics} \label{twelvetypes} 

In table \ref{tab:12types} below `$0$' flags the failure of a property and `$1$' means the property is verified. 

\begin{table}[h]
	 \caption{The $12$ possible types of $ASM^h$ and $ASM^m$ semantics}
	   	  \label{tab:12types}
     \begin{center}
		\begin{tabular}{cccccc}
          \hline \hline
       &$\exists$ & $gl$ & $lg$ & $cm$& $cut$ \\
       \hline
        $1$& $0$& $0$ & $0$& $0$ &$0$\\ 
        $2$& $0$& $0$ & $0$& $0$ &$1$\\ 
        $3$& $0$& $0$ & $1$& $0$ &$0$\\ 
        $4$& $0$& $0$ & $1$& $0$ &$1$\\ 
        $5$& $1$& $1$ & $0$& $0$ &$0$\\ 
        $6$& $1$& $1$ & $0$& $0$ &$1$\\ 
        $7$& $1$& $1$ & $0$& $1$ &$0$\\ 
        $8$& $1$& $1$ & $0$& $1$ &$1$\\ 
        $9$& $1$& $1$ & $1$& $0$ &$0$\\ 
        $10$& $1$& $1$ & $1$& $0$ &$1$\\ 
        $11$& $1$& $1$ & $1$& $1$ &$0$\\ 
        $12$& $1$& $1$ & $1$& $1$ &$1$\\ 
       	\hline
    \end{tabular}
\end{center}
\end{table}

The $20$ missing types of semantics correspond to cases where ($\exists=0$ and $gl=1$), or ($\exists=1$ and $gl=0$), or ($\exists=0$ and $cm=1$), each of these cases going against the statement of theorem \ref{teo:equivdefect}. The correspondence of the $ASM^h \cup ASM^m$ class semantics presented in this text and the entries in table \ref{tab:12types} is as follows: $1.$ $ MH^{sustainable}, MH^{Sustainable}_{min}$ $2.$ $ -- $ $3.$ $ -- $ $4.$ $ SM$ $5.$ $ MH, MH^{LS}, MH^{Loop}, Green$ $6.$ $ -- $ $7.$ $ Navy$ $8.$ $  Blue$ $9.$ $ MH^{Regular}$ $10.$ $ -- $ $11.$ $ -- $ $12.$ $ Cyan$. Whether semantics of the $ASM^h \cup ASM^m$ class exist for the types marked with '$ -- $', may be envisaged as an open issue.

\end{document}